\documentclass[letterpaper]{article} 
\usepackage{aaai2026}  
\usepackage{times}  
\usepackage{helvet}  
\usepackage{courier}  
\usepackage[hyphens]{url}  
\usepackage{graphicx} 
\usepackage{natbib}  
\usepackage{caption} 
\usepackage{algorithm}
\usepackage{algorithmic}
\usepackage{enumitem}
\usepackage{amsmath}
\usepackage[table,xcdraw]{xcolor}
\usepackage{adjustbox}
\usepackage{multirow}
\usepackage{newfloat}
\usepackage{listings}
\DeclareCaptionStyle{ruled}{labelfont=normalfont,labelsep=colon,strut=off} 
\floatstyle{ruled}
\newfloat{listing}{tb}{lst}{}
\floatname{listing}{Listing}
\title{Text-Guided Channel Perturbation and Pretrained Knowledge Integration for Unified Multi-Modality Image Fusion}
\author{
Xilai Li,
Xiaosong Li\thanks{Corresponding author},
Weijun Jiang
}

\affiliations{
College of Physics and Optoelectronic Engineering, Foshan University, Foshan 528225, China\\
20210300236@stu.fosu.edu.cn, lixiaosong@buaa.edu.cn, 20230300130@stu.fosu.edu.cn
}

\usepackage{bibentry}

\begin{document}

\maketitle

\begin{abstract}
Multi-modality image fusion enhances scene perception by combining complementary information. Unified models aim to share parameters across modalities for multi-modality image fusion, but large modality differences often cause gradient conflicts, limiting performance. Some methods introduce modality-specific encoders to enhance feature perception and improve fusion quality. However, this strategy reduces generalisation across different fusion tasks. To overcome this limitation, we propose a \textbf{U}nified multi-modality image fusion framework based on Channel \textbf{P}erturbation and Pre-trained Knowledge Integration (\textbf{UP-Fusion}). To suppress redundant modal information and emphasize key features, we propose the Semantic-Aware Channel Pruning Module (SCPM), which leverages the semantic perception capability of a pre-trained model to filter and enhance multi-modality feature channels. Furthermore, we proposed the Geometric Affine Modulation Module (GAM), which uses original modal features to apply affine transformations on initial fusion features to maintain the feature encoder modal discriminability. Finally, we apply a Text-Guided Channel Perturbation Module (TCPM) during decoding to reshape the channel distribution, reducing the dependence on modality-specific channels. Extensive experiments demonstrate that the proposed algorithm outperforms existing methods on both multi-modality image fusion and downstream tasks.
\end{abstract}


\begin{links}
    \link{Code}{https://github.com/ixilai/UP-Fusion}
\end{links}

\section{Introduction}

Multi-modality image fusion (MMIF) aims to combine complementary features from different imaging modalities to produce more comprehensive and informative representations \cite{r1,r2}. For example, infrared and visible image fusion (IVIF) \cite{r3,r4,r5,r52,r55} merges fine texture details from visible light with thermal radiation information from infrared imaging, improving the robustness of machine vision under complex lighting conditions. Similarly, medical image fusion (MEIF) \cite{r6,r7} integrates data from Computed Tomography (CT), Magnetic Resonance Imaging (MRI), Positron Emission Tomography (PET), and Single Photon Emission Computed Tomography (SPECT), enabling a more complete understanding of pathological features. Beyond these applications, MMIF also shows strong potential in downstream tasks such as semantic segmentation and object detection.
\begin{figure}[t]
  \centering
   \includegraphics[width=1\linewidth]{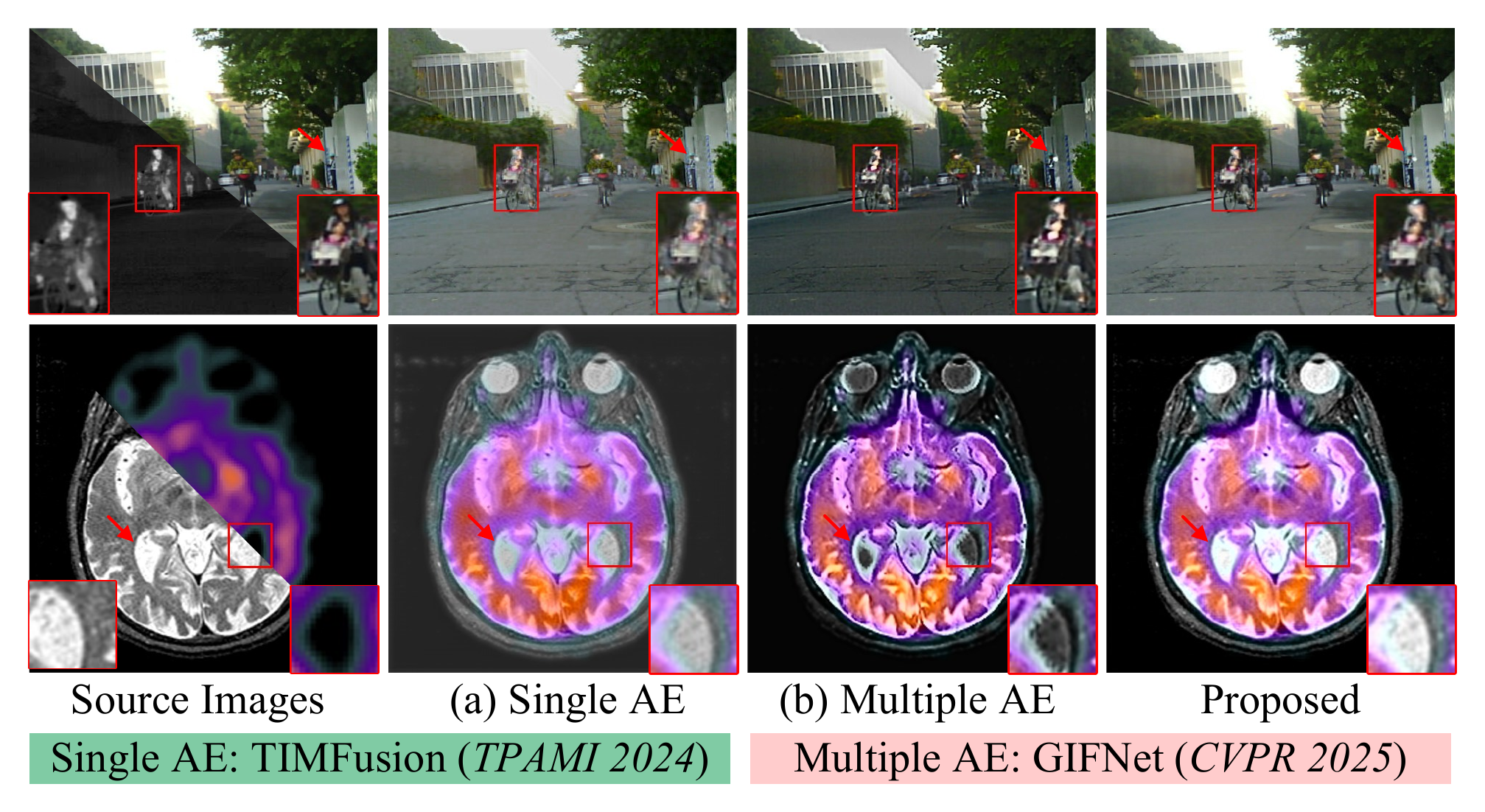}
   \caption{Comparison of the Single AE algorithm \cite{r14}, Multiple AE algorithm \cite{r37}, and the proposed method across different MMIF tasks.}
   \label{fig1}
\end{figure}

Recently, unified MMIF models have attracted increasing attention. A typical approach within this trend is based on a unified autoencoder (AE) structure \cite{r13,r15,r16}. In such methods, pre-fused features are typically obtained by channel-level concatenation or direct summation of multi-modal inputs, which are then passed through a shared encoder–decoder network for optimization. Alternatively, shared weights are employed to extract features from different modalities simultaneously. Some methods enhance adaptability to MMIF tasks by incorporating strategies such as continual learning \cite{r8} and meta-learning \cite{r14}. The single-encoder structure allows models to share feature extractors and decoder parameters across modalities, leading to strong cross-task generalisation. However, these approaches often lack explicit modeling of inter-modal interaction mechanisms, which limits their fusion performance. As shown in Fig.~\ref{fig1}, although the single-AE method can handle multiple MMIF tasks, its fusion quality remains suboptimal.
The second class of methods adopts modality-specific encoder architectures \cite{r17,r18,r19}. These approaches design separate encoders for each modality to extract modality-specific representations, which are then concatenated or fused before being processed by a unified decoder to generate the final fused image. This strategy effectively preserves the distinct feature expressiveness of each modality during the encoding stage. In addition, the method \cite{r42} introduces inter-modal interaction modules between encoders to enable cross-modal feature guidance, further improving fusion performance. However, modality-specific encoders have inherent limitations. Since each encoder tends to overfit the feature distribution of its corresponding modality during training, the model often suffers from poor generalization when applied to unseen modal combinations or datasets with different distributions. Consequently, many MMIF methods require training separate weights for different tasks rather than adopting a unified set of parameters. As shown in Fig.~\ref{fig1}, when a multi-AE algorithm trained on IVIF data is applied to the MEIF task, it tends to introduce erroneous background details instead of extracting salient and informative features.

To balance the generalization ability of a single AE with the superior fusion quality of modality-specific AEs, we propose a unified MMIF framework based on text-guided channel perturbation with pre-trained knowledge integration (UP-Fusion). Considering the high redundancy and weak complementarity in multi-modal data, and the tendency for feature extraction to cause substantial channel expansion, we design a Semantic-Aware Channel Pruning Module (SCPM). This module leverages a Squeeze-and-Excitation (SE) block to model channel response strength and integrates the global semantic awareness of the pre-trained ConvNeXt model, jointly guiding the identification of salient channels. Subsequently, to enhance the representation and fusion of modality-specific features, we propose the Geometric Affine Modulation Module (GAM). This module performs structural modulation of pre-fused features through affine transformations. In the decoding phase, we propose the Text-Guided Channel Perturbation Module (TCPM), which utilises semantic text to guide the channel selection and alignment of multi-modality features after first using channel attention for further filtering, enabling the features to remove modal markers and improving the generalisation ability of the model. Our main contributions are as follows:

\begin{itemize}[label=$\bullet$]
\item We propose a unified multi-modality image fusion framework based on text-guided channel perturbation and pre-trained knowledge, effectively reducing modal redundancy while enhancing unified modeling and generalization across cross-modal features.

\item We propose the Semantic-Aware Channel Pruning Module, which integrates semantic awareness with channel attention to filter redundant channels and strengthen salient features. Additionally, we propose the Geometric Affine Modulation Module to perform affine transformations on fusion representations based on modality-specific features.

\item Extensive experiments demonstrate that the proposed method significantly outperforms existing task-specific and unified fusion models in both infrared-visible and medical image fusion tasks, while also exhibiting superior performance in downstream applications.

\end{itemize}

\section{Related Works}
\subsection{Multi-Modality Image Fusion}
Recent advances in MMIF have primarily centered on deep learning-based frameworks \cite{r20,r21,r57}. Mainstream approaches can be broadly divided into two categories: autoencoder (AE)-based methods and generative model-based methods. The first group includes models built upon Convolutional Neural Networks (CNNs) \cite{r8}, Transformers \cite{r9}, Mamba architectures \cite{r10}, and their hybrid variants \cite{r4}, whereas the second group primarily relies on Generative Adversarial Networks (GANs) \cite{r11} and diffusion models \cite{r12}. For example, \citet{r17} proposed a coupled contrastive learning network for MMIF. This method effectively preserves salient information from different modalities and reduces redundancy by incorporating a contrastive constraint mechanism into the loss function. \citet{r19} proposed an equivariant MMIF framework that integrates the equivariant prior of natural images into the training process to improve fusion performance. To address the instability of GAN-based methods, \citet{r12} proposed an MMIF algorithm using the Denoising Diffusion Probabilistic Model, framing the fusion task as a conditional generation problem.

In addition, some researchers have extended the MMIF technique to downstream tasks \cite{r22,r41}(e.g., detection and segmentation) and complex scenarios \cite{r23,r24,r25,r54,r56} (e.g., adverse weather and low light).
Specifically, \citet{r26} distilled semantic knowledge from the Segment Anything Model into the fusion network, leveraging high-level semantic priors to guide the fusion process. Similarly, \citet{r18} proposed a learnable task-guided fusion loss, enabling the fusion results to adapt effectively to various downstream tasks. Meanwhile, some studies focus on improving the generalisation and applicability of fusion methods in complex scenes. For example, \citet{r23} proposed a MMIF method for adverse weather by designing an integrated restoration–fusion model, enabling feature restoration and interaction simultaneously. \citet{r25} introduced a degradation-aware interactive fusion network that leverages text guidance to enhance fusion performance. 

Despite recent progress in MMIF and its application in downstream tasks and complex scenarios, the design of unified model architectures remains overlooked. Existing methods typically rely on learning strategies to unify weights, such as U2Fusion \cite{r8} with continual learning and GIFNet \cite{r37} with task-specific constraints. \textbf{In contrast, we explore the construction of unified models from the perspectives of single and modality-specific AEs, offering new insights to advance the field.}
\begin{figure*}[t]
  \centering
   \includegraphics[width=1\linewidth]{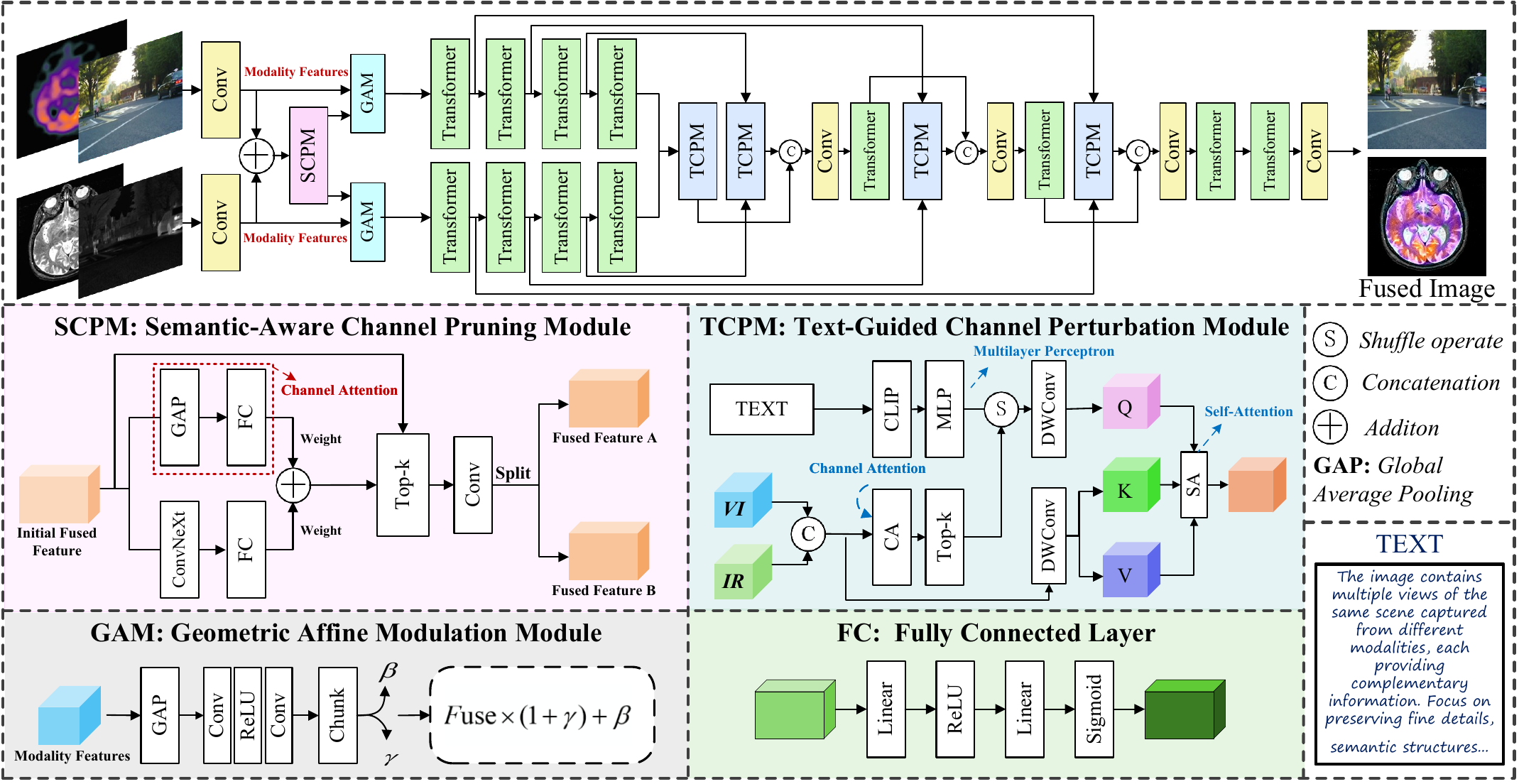}
   \caption{The overall framework of the proposed unified multi-modality image fusion framework.}
   \label{fig2}
\end{figure*}

\section{Proposed Method}
\subsection{Overview}
As shown in Fig.~\ref{fig2}, the proposed algorithm adopts an AE-based architecture built upon Transformer blocks \cite{r27}. The network comprises a 4-layer encoder and a 4-layer decoder, where the number of Transformer blocks in the encoder progressively increases as $[4, 6, 6, 8]$, along with a corresponding rise in attention heads ($[1, 2, 4, 8]$), enabling gradual spatial resolution reduction while expanding channel capacity.
In the encoder stage, multi-modal inputs $I^A$ and $I^B$ are first processed through convolutional layers. The proposed SCPM is then applied to remove redundant channels while retaining salient ones, followed by channel expansion to restore the original dimensionality. To incorporate modality-specific cues, we introduce the GAM module, which performs affine transformations on the pre-fused features under the guidance of the original modal representations. The dual-branch encoder extracts the modulated multi-modality features independently and performs channel-level interaction using the proposed TCPM at the deepest layer. The decoder, composed of four Transformer blocks, progressively reconstructs the fused image. At each stage, the corresponding encoder features are integrated and refined through TCPM, then concatenated with the upsampled features. Finally, channel compression via convolution ensures stepwise reconstruction of the fused image.

\subsection{Semantic-Aware Channel Pruning Module}
In MMIF tasks, redundant features often cause modal bias or feature interference. This redundancy suppresses complementary information, ultimately diminishing the semantic expressiveness and visual fidelity of the fused image. To address this, we propose the Semantic-Aware Channel Pruning Module (SCPM), which adaptively preserves the most representative salient features for subsequent encoding by integrating a channel attention mechanism with global semantic guidance.

Specifically, we first compute the channel importance weight $\omega_C$ using the channel attention mechanism. Then, the pre-trained ConvNeXt backbone \cite{r28} is employed to extract semantic features from the initial fused representation, producing semantic weight $\omega_S$ through a linear mapping. The final channel weight $\omega_F$ is obtained by weighted fusion of $\omega_C$ and $\omega_S$,
\begin{equation}
\omega_F = \omega_C + \alpha \cdot \sigma(\omega_S)
\label{eq1}
\end{equation}
where $\alpha$ is a learnable parameter for balancing the weights and $\sigma$ is the Sigmoid activation function. According to $\omega_F$, SCPM adopts the Top-$k$ strategy to retain the top $70\%$ of the significant channels, and expands back to the original channel dimensions by $1 \times 1$ convolution. Finally, the output features are divided into two subsets ${Fuse}^{M}, \; M \in [A,B]$ by channel partitioning operation, which are sent to GAM for further processing.

\subsection{Geometric Affine Modulation Module}
Existing bimodal encoder-based approaches typically introduce modality-specific features directly for feature extraction and interaction. Although this strategy improves fusion performance, it tends to produce modality-dependent algorithms, as the encoder overfits to modality-specific features and thus generalizes poorly to unseen MMIF tasks. To overcome this limitation, we propose the GAM, which modulates fused features from a geometric perspective to mitigate the dependency of modality.

Specifically, we first perform global average pooling on the original multi-modality features to obtain $F_G^A$ and $F_G^B$, respectively. Subsequently, the mapping is performed by a fully-connected network of two-layer $1 \times 1$ convolutions to obtain the scaling factor $\gamma$ and the offset term $\beta$ of the affine transformation:
\begin{equation}
[\gamma^{M}, \beta^{M}] = \mathrm{Conv}_{1\times 1}\big(\mathrm{ReLU}(\mathrm{Conv}_{1\times 1}(F_G^{M}))\big)
\label{eq2}
\end{equation}
Then, we perform affine modulation of the fused features:
\begin{equation}
F_O^{M} = {Fuse}^{M} \cdot (1 + \gamma) + \beta
\label{eq3}
\end{equation}
By modulating with two parameters—translation and scaling—GAM enables the fusion features refined by SCPM to adjust their spatial distribution according to the global geometric characteristics of the original modality.

\subsection{Text-Guided Channel Perturbation Module}
After obtaining feature representations of different modalities through a multi-modality encoder, existing methods typically perform a weighted combination and input the result into the decoding module for reconstruction. This approach offers simplicity and fast convergence on modality-specific tasks. However, because this operation is inherently unlearnable, it relies heavily on the modality-specific data distribution during training. Consequently, the decoding process cannot adapt dynamically when the model is applied to different multimodal image fusion tasks. To address this limitation, inspired by \citet{r29}, we introduce the concept of channel perturbation into MMIF and propose the TCPM. TCPM uses channel selection and rearrangement, guided by textual semantics, to perturb modal feature distribution during decoding.

Specifically, multi-modality features are first concatenated along the channel dimension, and the importance of each channel is evaluated using a channel attention mechanism. Then, a Top-$k$ strategy selects the top $50\%$ most significant channels. To alleviate gradient disturbance caused by channel rearrangement, a $1 \times 1$ convolution is applied to expand the channel number to twice the original, enhancing feature redundancy and expressiveness. Simultaneously, to realize text-guided channel perturbation, CLIP-encoded \cite{r30} text features are passed through a two-layer linear mapping network to produce bootstrap weights matching the channel dimension. These weights are used to generate a channel rearrangement index \cite{r29}, guiding the fused features to undergo semantically relevant channel rearrangement. To further enhance interaction between perturbed and original features, a self-attention mechanism \cite{r29} is employed, where the perturbed features serve as the query ($Q$), and the original features provide the key ($K$) and value ($V$). The resulting output is processed through a feed-forward network.

\subsection{Loss Function}
During training, we used two loss functions: gradient loss and L1 loss, to preserve image details and structural information. The gradient loss is defined as:
\begin{equation}
L_{\text{grad}} = \frac{1}{HW} \left\| \nabla {Fuse} - \max\big( \nabla {I_A}, \nabla {I_B} \big) \right\|_{1}
\label{eq4}
\end{equation}
where $H$ and $W$ are the length and width of the image, respectively, $Fuse$ is the final fusion result, and $I_A$ and $I_B$ are the multi-modality source images, respectively. The L1 loss is calculated as follows.
\begin{equation}
L_{l_1} = \frac{1}{HW} \sum_{M \in \{A,B\}} \left\| {Fuse} - {I_M} \right\|_{1}
\label{eq5}
\end{equation}
The total loss $L_{\text{T}}$ can be expressed as $L_{\text{T}} = L_{\text{grad}} + L_{l_1}$
\begin{figure}[t]
  \centering
   \includegraphics[width=1\linewidth]{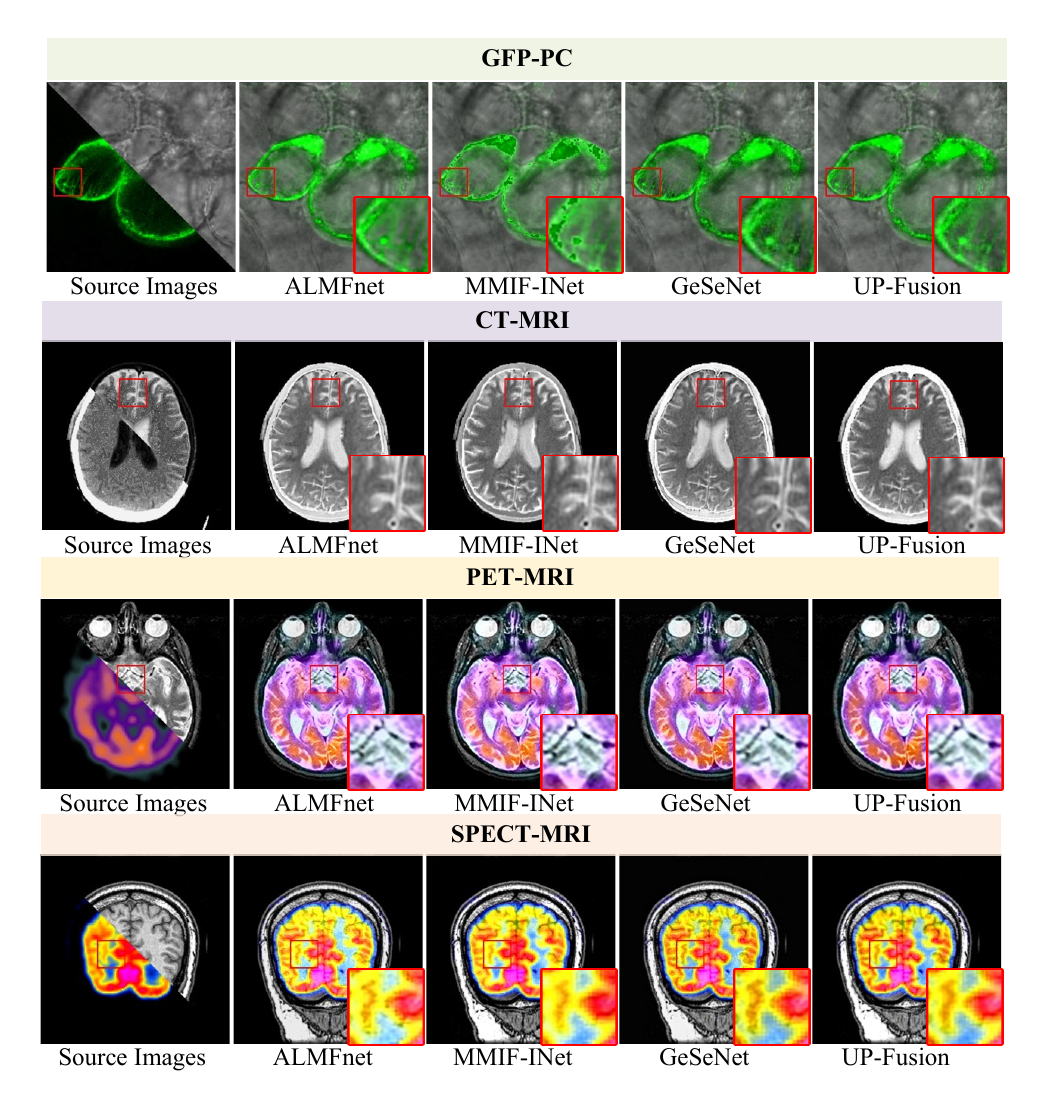}
   \caption{Comparison of the proposed algorithm and medical image fusion methods on medical image tasks.}
   \label{fig4}
\end{figure}
\begin{figure*}[t]
  \centering
   \includegraphics[width=1\linewidth]{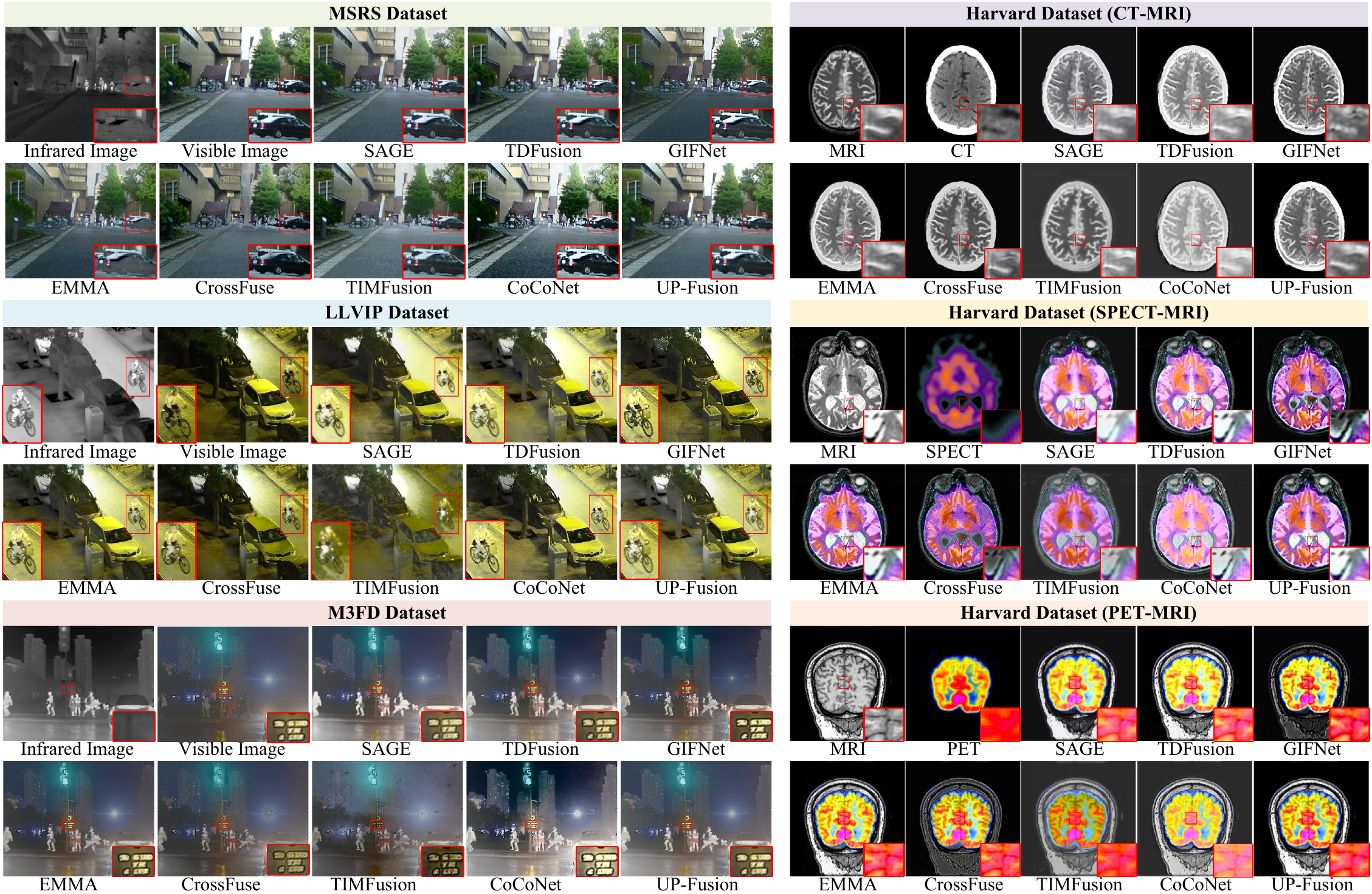}
   \caption{Comparison of different methods on infrared and visible image fusion, and medical image fusion tasks.}
   \label{fig3}
\end{figure*}
\section{Experiments}
\subsection{Experimental Setup Details}
\subsubsection{Dataset} 
To evaluate the generality of the proposed method, we conducted experiments on two tasks: infrared-visible image fusion and medical image fusion. For the former, three mainstream datasets were used: MSRS \cite{r31}, LLVIP \cite{r32}, and M3FD \cite{r11}. For the latter, we used multi-modality brain medical images from Harvard Medical School\footnote{http://www.med.harvard.edu/aanlib/home.html}, covering three typical fusion scenarios: CT-MRI, PET-MRI, and SPECT-MRI. In addition, we tested generalisation on a mainstream database\footnote{http://data.jic.ac.uk/Gfp/} containing green fluorescent protein (GFP) and phase-contrast (PC) images.
\subsubsection{Metrics} 
We used five metrics to objectively evaluate fusion performance: Nonlinear Correlation Information Entropy ($Q_{NCIE}$), Phase Congruency-based Image Fusion Metric ($Q_P$), Visual Information Fidelity ($VIF$), Structural Similarity Index Measure ($SSIM$), and Normalized Weighted Performance Metric ($Q^{AB/F}$) \cite{r38}.
\subsubsection{Comparison methods} 
We compared with 10 state-of-the-art MMIF methods, including three IVIF methods (CrossFuse \cite{r33}, SAGE \cite{r26}, TDFusion \cite{r18}), three MEIF methods (ALMFNet \cite{r34}, MMIF-INet \cite{r35}, GeSeNet \cite{r36}), and four general MMIF methods (GIFNet \cite{r37}, EMMA \cite{r19}, TIMFusion \cite{r14}, CoCoNet \cite{r17}).
\subsubsection{Training Details} 
The proposed method is trained only on the LLVIP dataset and uses a single weight for all MMIF tasks. We conducted training for a total of 100 epochs. The images were randomly cropped to a size of $192 \times 192$ pixels, with a batch size of $2$. The initial learning rate was set to $0.0001$ and gradually decayed to $0.00001$ using a cosine decay schedule. The optimization was performed using the Adam optimizer. All experiments are conducted in a PyTorch 2.1.1 environment on a server equipped with four GeForce RTX 3090 GPUs.

\subsection{Qualitative Comparison}
Fig.~\ref{fig4} shows the fusion results of the proposed algorithm (UP-Fusion) compared with existing MEIF methods on brain images from Harvard Medical School and GFP-PC images. Although trained only on IVIF data, our method achieves excellent performance on medical image tasks, accurately extracting bone structures in CT images and detailed textures in MRI images. It also outperforms medical-specific fusion methods in color fidelity. These results indicate that the proposed channel perturbation mechanism effectively reduces modality dependence and significantly enhances the cross-task ability of the proposed method.

Fig.~\ref{fig3} presents the visual results of the proposed algorithm compared with three IVIF methods and four general MMIF algorithms on both IVIF and MEIF tasks. Methods such as SAGE, TDFusion, GIFNet, EMMA, CrossFuse, and CoCoNet, like ours, are trained only on infrared-visible images and generally perform well on IVIF tasks. However, when applied to medical image fusion, SAGE and CoCoNet lose detail, while GIFNet and CrossFuse introduce excessive black background regions, degrading fusion quality. Although TDFusion and EMMA remain stable on medical images, they exhibit contrast loss on IVIF tasks. In contrast, the proposed method achieves excellent performance across multiple IVIF datasets and medical fusion tasks.

\begin{table*}[t]
\begin{adjustbox}{width=\textwidth} 
\begin{tabular}{cc|ccccc|ccccc}
\hline
\multicolumn{2}{c|}{Datasets}                     & \multicolumn{5}{c|}{MSRS}                                                                                                                                          & \multicolumn{5}{c}{LLVIP}                                                                                                                                          \\ \hline
\multicolumn{1}{c|}{Methods}   & Pub.             & \textbf{$Q_{NCIE}$↑}                 & \textbf{$Q_{P}$↑}                    & \textbf{$VIF$↑}                   & \textbf{$SSIM$↑}                  & \textbf{$Q^{AB/F}$↑}                 & \textbf{$Q_{NCIE}$↑}                 & \textbf{$Q_P$↑}                    & \textbf{$VIF$↑}                   & \textbf{$SSIM$↑}                  & \textbf{$Q^{AB/F}$↑}                 \\ \hline
\multicolumn{1}{c|}{SAGE}      & \textit{CVPR 25} & 0.8123                         & 0.5210                         & \cellcolor[HTML]{D9E1F4}0.4359 & 0.4691                         & 0.6242                         & 0.8065                         & 0.4128                         & \cellcolor[HTML]{D9E1F4}0.3590 & 0.4223                         & 0.5631                         \\
\multicolumn{1}{c|}{TDFusion}  & \textit{CVPR 25} & \cellcolor[HTML]{D9E1F4}0.8159 & \cellcolor[HTML]{D9E1F4}0.5529 & 0.4257                         & \cellcolor[HTML]{D9E1F4}0.5010 & \cellcolor[HTML]{D9E1F4}0.6770 & 0.8082                         & \cellcolor[HTML]{D9E1F4}0.4406 & 0.3577                         & \cellcolor[HTML]{D9E1F4}0.4485 & \cellcolor[HTML]{D9E1F4}0.6252 \\
\multicolumn{1}{c|}{GIFNet}    & \textit{CVPR 25} & 0.8048                         & 0.3314                         & 0.2781                         & 0.4192                         & 0.4542                         & 0.8058                         & 0.3399                         & 0.2126                         & 0.3855                         & 0.4515                         \\
\multicolumn{1}{c|}{EMMA}      & \textit{CVPR 24} & 0.8110                         & 0.4407                         & 0.3689                         & 0.4613                         & 0.5947                         & 0.8075                         & 0.4183                         & 0.3301                         & 0.4339                         & 0.5969                         \\
\multicolumn{1}{c|}{CrossFuse} & \textit{INF 24}  & 0.8142                         & 0.3564                         & 0.3246                         & 0.3931                         & 0.4943                         & \cellcolor[HTML]{FADADE}0.8129 & 0.4095                         & 0.3398                         & 0.4396                         & 0.6246                         \\
\multicolumn{1}{c|}{TIMFusion}       & \textit{PAMI 24} & 0.8051                         & 0.1913                         & 0.1834                         & 0.2834                         & 0.4078                         & 0.8042                         & 0.1361                         & 0.2018                         & 0.2629                         & 0.2754                         \\
\multicolumn{1}{c|}{CoCoNet}   & \textit{IJCV 24} & 0.8061                         & 0.2925                         & 0.1669                         & 0.2987                         & 0.3547                         & 0.8062                         & 0.4007                         & 0.2434                         & 0.3799                         & 0.5081                         \\
\multicolumn{1}{c|}{UP-Fusion}  & w/o              & \cellcolor[HTML]{FADADE}0.8167 & \cellcolor[HTML]{FADADE}0.5671 & \cellcolor[HTML]{FADADE}0.4587 & \cellcolor[HTML]{FADADE}0.5074 & \cellcolor[HTML]{FADADE}0.6859 & \cellcolor[HTML]{D9E1F4}0.8092 & \cellcolor[HTML]{FADADE}0.5042 & \cellcolor[HTML]{FADADE}0.3817 & \cellcolor[HTML]{FADADE}0.4636 & \cellcolor[HTML]{FADADE}0.6864 \\ \hline
\multicolumn{2}{c|}{Datasets}                     & \multicolumn{5}{c|}{M3FD}                                                                                                                                          & \multicolumn{5}{c}{Harvard}                                                                                                                                        \\ \hline
\multicolumn{1}{c|}{Methods}   & Pub.             & \textbf{$Q_{NCIE}$↑}                 & \textbf{$Q_P$↑}                    & \textbf{$VIF$↑}                   & \textbf{$SSIM$↑}                  & \textbf{$Q^{AB/F}$↑}                 & \textbf{$Q_{NCIE}$↑}                 & \textbf{$Q_P$↑}                    & \textbf{$VIF$↑}                   & \textbf{$SSIM$↑}                  & \textbf{$Q^{AB/F}$↑}                 \\ \hline
\multicolumn{1}{c|}{SAGE}      & \textit{CVPR 25} & 0.8065                         & 0.4813                         & \cellcolor[HTML]{D9E1F4}0.4110 & 0.4909                         & 0.5974                         & 0.8058                         & 0.3374                         & 0.2263                         & 0.2054                         & 0.4244                         \\
\multicolumn{1}{c|}{TDFusion}  & \textit{CVPR 25} & 0.8065                         & 0.4748                         & 0.4041                         & \cellcolor[HTML]{FADADE}0.5231 & \cellcolor[HTML]{D9E1F4}0.6270 & 0.8070                         & 0.4961                         & \cellcolor[HTML]{D9E1F4}0.2859 & \cellcolor[HTML]{FADADE}0.4518 & 0.6410                         \\
\multicolumn{1}{c|}{GIFNet}    & \textit{CVPR 25} & 0.8061                         & 0.4002                         & 0.2918                         & 0.4710                         & 0.5243                         & 0.8054                         & 0.2784                         & 0.1689                         & 0.1942                         & 0.4195                         \\
\multicolumn{1}{c|}{EMMA}      & \textit{CVPR 24} & 0.8109                         & \cellcolor[HTML]{FADADE}0.5341 & 0.3829                         & 0.4787                         & 0.6032                         & 0.8068                         & \cellcolor[HTML]{D9E1F4}0.5154 & 0.2745                         & 0.3029                         & \cellcolor[HTML]{D9E1F4}0.6489 \\
\multicolumn{1}{c|}{CrossFuse} & \textit{INF 24}  & \cellcolor[HTML]{FADADE}0.8140 & 0.4122                         & 0.3807                         & 0.4634                         & 0.5844                         & \cellcolor[HTML]{D9E1F4}0.8072 & 0.3626                         & 0.2390                         & 0.2232                         & 0.5645                         \\
\multicolumn{1}{c|}{TIMFusion}       & \textit{PAMI 24} & 0.8055                         & 0.2926                         & 0.2680                         & 0.3928                         & 0.4912                         & 0.8056                         & 0.3671                         & 0.2299                         & 0.1600                         & 0.3364                         \\
\multicolumn{1}{c|}{CoCoNet}   & \textit{IJCV 24} & 0.8049                         & 0.3242                         & 0.1967                         & 0.3288                         & 0.3447                         & 0.8056                         & 0.3900                         & 0.2068                         & 0.1872                         & 0.4750                         \\
\multicolumn{1}{c|}{UP-Fusion}  & w/o              & \cellcolor[HTML]{D9E1F4}0.8119 & \cellcolor[HTML]{D9E1F4}0.5190 & \cellcolor[HTML]{FADADE}0.4582 & \cellcolor[HTML]{D9E1F4}0.5163 & \cellcolor[HTML]{FADADE}0.6275 & \cellcolor[HTML]{FADADE}0.8074 & \cellcolor[HTML]{FADADE}0.5665 & \cellcolor[HTML]{FADADE}0.3190 & \cellcolor[HTML]{D9E1F4}0.3639 & \cellcolor[HTML]{FADADE}0.6814 \\ \hline
\end{tabular}
\end{adjustbox}
\caption{Quantitative comparison of different methods across multiple multi-modality image fusion datasets. The marked red indicates the best score, and the marked blue indicates the second score.}
\label{tab1}
\end{table*}
\begin{table*}[h]
\begin{adjustbox}{width=\textwidth} 
\begin{tabular}{cc|ccccc|ccccc}
\hline
\multicolumn{2}{c|}{Datasets}                        & \multicolumn{5}{c|}{GFP-PC}                                                                                                                                        & \multicolumn{5}{c}{Harvard}                                                                                                                                        \\ \hline
\multicolumn{1}{c|}{Methods}   & Pub.             & \textbf{$Q_{NCIE}$↑}                 & \textbf{$Q_{P}$↑}                    & \textbf{$VIF$↑}                   & \textbf{$SSIM$↑}                  & \textbf{$Q^{AB/F}$↑}                 & \textbf{$Q_{NCIE}$↑}                 & \textbf{$Q_P$↑}                    & \textbf{$VIF$↑}                   & \textbf{$SSIM$↑}                  & \textbf{$Q^{AB/F}$↑}                 \\ \hline
\multicolumn{1}{c|}{ALMFnet}   & \textit{TCSVT 23} & 0.8102                         & \cellcolor[HTML]{D9E1F4}0.4813                         & \cellcolor[HTML]{D9E1F4}0.3077                         & 0.3140                         & 0.5839                         & 0.8068                         & 0.5434                         & \cellcolor[HTML]{D9E1F4}0.3003 & \cellcolor[HTML]{FADADE}0.5777 & 0.6780                         \\
\multicolumn{1}{c|}{GeSeNet}   & \textit{TNNLS 23}   & \cellcolor[HTML]{D9E1F4}0.8146                         & 0.4740                         & 0.3049                         & \cellcolor[HTML]{D9E1F4}0.3202                         & \cellcolor[HTML]{FADADE}0.6044 & 0.8066                         & 0.5576                         & 0.2900                         & \cellcolor[HTML]{D9E1F4}0.5425 & \cellcolor[HTML]{D9E1F4}0.7058 \\
\multicolumn{1}{c|}{MMIF-INet} & \textit{INF 25}     & 0.8078                         & 0.3274                         & 0.2448                         & 0.2966                         & 0.4949                         & \cellcolor[HTML]{D9E1F4}0.8071 & \cellcolor[HTML]{D9E1F4}0.5591 & 0.2958                         & 0.2599                         & \cellcolor[HTML]{FADADE}0.7155 \\
\multicolumn{1}{c|}{UP-Fusion}  & w/o                 & \cellcolor[HTML]{FADADE}0.8170 & \cellcolor[HTML]{FADADE}0.4826 & \cellcolor[HTML]{FADADE}0.3193 & \cellcolor[HTML]{FADADE}0.3339 & \cellcolor[HTML]{D9E1F4}0.5914 & \cellcolor[HTML]{FADADE}0.8074 & \cellcolor[HTML]{FADADE}0.5665 & \cellcolor[HTML]{FADADE}0.3190 & 0.3639                         & 0.6814                         \\ \hline
\end{tabular}
\end{adjustbox}
\caption{Quantitative comparison of medical image fusion methods on a medical image dataset. The marked red indicates the best score, and the marked blue indicates the second score.}
\label{tab2}
\end{table*}
\begin{table}[h]
\footnotesize
\begin{tabular}{c|cccc}
\hline
Methods       & \textbf{$Q_{NCIE}$↑}                 & \textbf{$Q_P$↑}                    & \textbf{$VIF$↑}                   & \textbf{$SSIM$↑}                                 \\ \hline
w/o SCPM      & 0.8052                         & 0.5343                         & 0.3046                         & 0.2645                                      \\
w/o GAM       & 0.8071                         & 0.5488                         & 0.3151                         & 0.3478                                           \\
w/o TCPM      & 0.8054                         & 0.5221                         & 0.3016                         & 0.2824                                        \\
w/o CA (TCPM) & 0.8068                         & 0.5057                         & 0.2983                         & 0.2632                                            \\
w/o ConvNeXt  & 0.8072                         & 0.5478                         & 0.3018                         & 0.2807                                               \\
w/o CA (SCPM) & 0.8071                         & 0.5441                         & 0.3052                         & 0.3218                                           \\
UP-Fusion      & \cellcolor[HTML]{FADADE}0.8074 & \cellcolor[HTML]{FADADE}0.5665 & \cellcolor[HTML]{FADADE}0.3190 & \cellcolor[HTML]{FADADE}0.3639  \\ \hline
\end{tabular}
\caption{Quantitative comparison of different ablation results. The marked red indicates the best score.}
\label{tab3}
\end{table}

\subsection{Quantitative Comparison}
Tabs.~\ref{tab1} and ~\ref{tab2} present the quantitative comparison results of the proposed algorithm on IVIF and MEIF tasks, respectively. Table 1 evaluates the performance of the proposed method against IVIF and general MMIF algorithms on IVIF datasets and examines its generalisation to MEIF tasks. Table 2 compares the proposed algorithm with MEIF-specific methods on MEIF tasks, including GFP-PC images.

As shown in Tab.~\ref{tab1}, the proposed algorithm achieves the optimum in multiple metrics on the MSRS, LLVIP, and M3FD datasets. In the MEIF task, it also ranks first in four metrics and second in one, showing superior cross-modal generalisation ability. Tab.~\ref{tab2} further shows that the proposed algorithm outperforms fusion methods designed for medical images in several metrics.

\begin{figure}[t]
  \centering
   \includegraphics[width=1\linewidth]{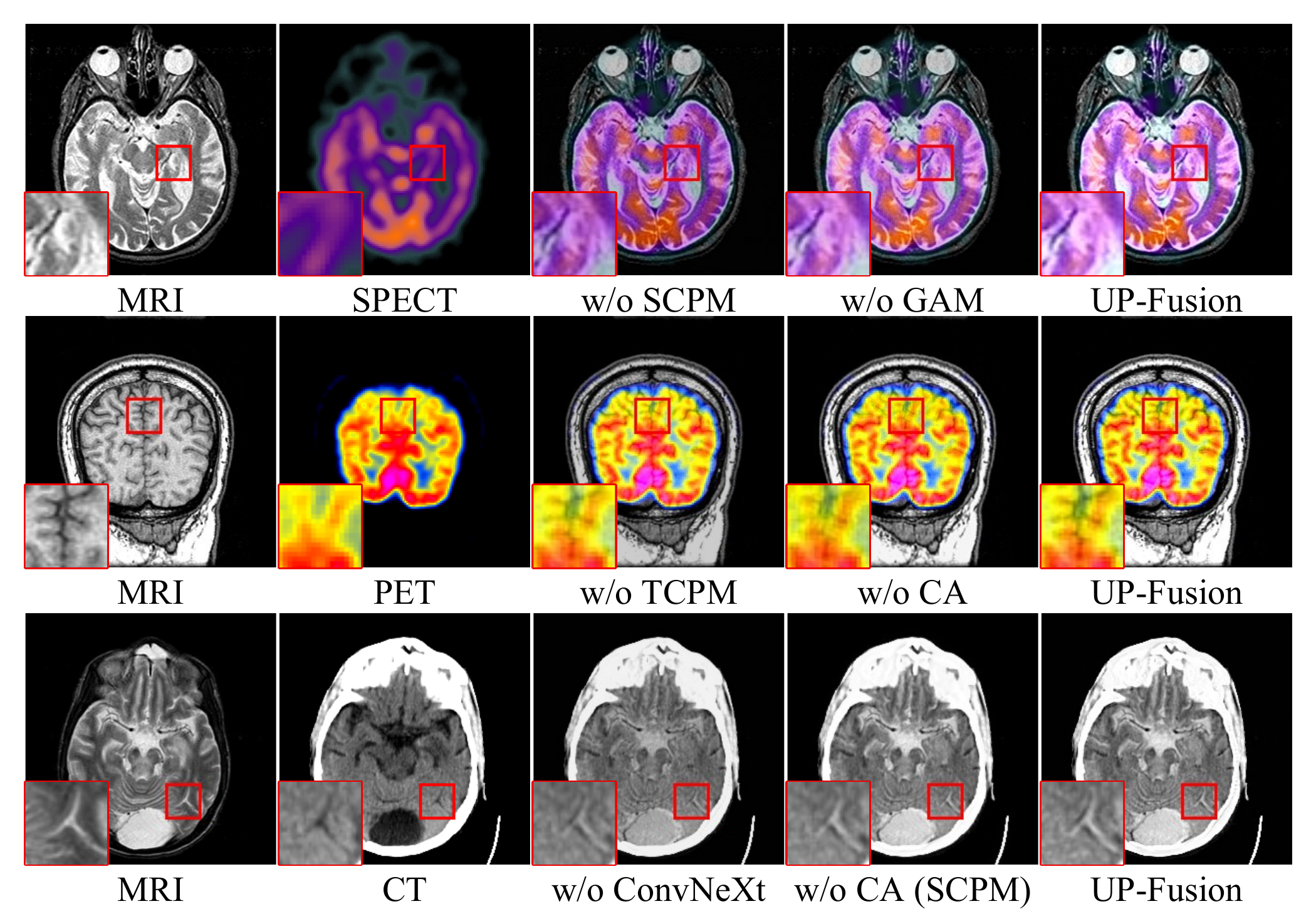}
   \caption{Different ablation results on medical image fusion.}
   \label{fig5}
\end{figure}

\begin{table*}[t]
\begin{adjustbox}{width=\textwidth}
\begin{tabular}{c|c|cccccccccc}
\hline
Methods   & Pub.             & Background                    & Car                           & Person                        & Bike                          & Curve                         & Car Stop                      & Cuardrail                     & Color cone                    & Bump                          & mIoU                          \\ \hline
SAGE      & \textit{CVPR 25} & 98.38                         & 88.8                          & 69.53                         & 70.37                         & 57.13                         & 71.21                         & \cellcolor[HTML]{FADADE}86.06 & 65.08                         & 76.9                          & 75.94                         \\
TDFusion  & \textit{CVPR 25} & 98.51                         & \cellcolor[HTML]{D9E1F4}90.37 & \cellcolor[HTML]{D9E1F4}74.43 & 71.9                          & \cellcolor[HTML]{FADADE}65.02 & \cellcolor[HTML]{D9E1F4}74.39 & 83.87                         & \cellcolor[HTML]{D9E1F4}65.94 & \cellcolor[HTML]{D9E1F4}78.85 & \cellcolor[HTML]{D9E1F4}78.14 \\
GIFNet    & \textit{CVPR 25} & 98.4                          & 88.98                         & 71.13                         & \cellcolor[HTML]{D9E1F4}71.92 & 59.19                         & 74.11                         & 78.05                         & 65.3                          & 64.62                         & 74.63                         \\
EMMA      & \textit{CVPR 24} & \cellcolor[HTML]{D9E1F4}98.53 & 90.36                         & \cellcolor[HTML]{FADADE}74.65 & 71.61                         & \cellcolor[HTML]{D9E1F4}64.61 & \cellcolor[HTML]{FADADE}74.49 & 83.73                         & 65.61                         & 76.36                         & 77.77                         \\
CrossFuse & \textit{INF 24}  & 98.11                         & 86.89                         & 65.17                         & 69.65                         & 49.31                         & 69.85                         & 75.76                         & 65.56                         & 59.36                         & 70.74                         \\
TIMFuison       & \textit{PAMI 24} & 97.76                         & 82.87                         & 66.34                         & 60.74                         & 40.07                         & 64.48                         & 77.93                         & 54.77                         & 55.1                          & 66.67                         \\
CoCoNet   & \textit{IJCV 24} & 98.12                         & 85.83                         & 68.66                         & 70.47                         & 52.57                         & 70.05                         & 74.03                         & 60.89                         & 53.9                          & 70.5                          \\
UP-Fusion  & w/o              & \cellcolor[HTML]{FADADE}98.56 & \cellcolor[HTML]{FADADE}90.49 & 73.47                         & \cellcolor[HTML]{FADADE}71.94 & 64.02                         & 74.05                         & \cellcolor[HTML]{D9E1F4}85.63 & \cellcolor[HTML]{FADADE}66.1  & \cellcolor[HTML]{FADADE}80.22 & \cellcolor[HTML]{FADADE}78.28 \\ \hline
\end{tabular}
\end{adjustbox}
\caption{Quantitative comparison of segmentation accuracy for different algorithms on the MSRS dataset.}
\label{tab4}
\end{table*}

\begin{table*}[t]
\centering
\begin{tabular}{c|c|cccccccc}
\hline
Methods   & Pub.             & People                        & Car                           & Bus                           & Lamp                          & Motorcycle                    & Truck                         & mAP@0.5                       &  mAP@{[}0.5:0.95{]} \\ \hline
SAGE      & \textit{CVPR 25} & 0.823                         & 0.918                         & \cellcolor[HTML]{D9E1F4}0.923 & \cellcolor[HTML]{D9E1F4}0.817 & 0.704                         & \cellcolor[HTML]{FADADE}0.832 & \cellcolor[HTML]{D9E1F4}0.836 & 0.532                                           \\
TDFusion  & \textit{CVPR 25} & \cellcolor[HTML]{FADADE}0.831 & \cellcolor[HTML]{D9E1F4}0.919 & 0.919                         & 0.806                         & \cellcolor[HTML]{D9E1F4}0.715 & \cellcolor[HTML]{D9E1F4}0.826 & 0.836                         & \cellcolor[HTML]{D9E1F4}0.535                   \\
GIFNet    & \textit{CVPR 25} & 0.805                         & 0.907                         & 0.919                         & 0.785                         & 0.685                         & 0.783                         & 0.814                         & 0.506                                           \\
EMMA      & \textit{CVPR 24} & 0.819                         & 0.9                           & 0.891                         & 0.727                         & 0.649                         & 0.787                         & 0.796                         & 0.498                                           \\
CrossFuse & \textit{INF 24}  & 0.716                         & 0.907                         & 0.901                         & 0.792                         & 0.699                         & 0.807                         & 0.804                         & 0.503                                           \\
TIMFuison       & \textit{PAMI 24} & 0.733                         & 0.906                         & 0.907                         & 0.783                         & 0.656                         & 0.779                         & 0.794                         & 0.499                                           \\
CoCoNet   & \textit{IJCV 24} & 0.764                         & 0.899                         & 0.913                         & 0.748                         & 0.664                         & 0.743                         & 0.789                         & 0.485                                           \\
UP-Fusion  & w/o              & \cellcolor[HTML]{D9E1F4}0.825 & \cellcolor[HTML]{FADADE}0.922 & \cellcolor[HTML]{FADADE}0.927 & \cellcolor[HTML]{FADADE}0.823 & \cellcolor[HTML]{FADADE}0.728 & 0.819                         & \cellcolor[HTML]{FADADE}0.841 & \cellcolor[HTML]{FADADE}0.541                   \\ \hline
\end{tabular}
\caption{Quantitative comparison of detection accuracy for different algorithms on the M3FD dataset.}
\label{tab5}
\end{table*}

\subsection{Ablation studies}
Ablation studies were performed to evaluate each module, with experiments carried out primarily on the MEIF task.

\subsubsection{Analysis of SCPM}
SCPM suppresses redundant features and reduces modality dependence in the encoder. As shown in Fig.~\ref{fig5}, removing SCPM markedly degrades contrast and fine structural details in medical images. Correspondingly, information and structure–preservation metrics (e.g., $Q_{NCIE}$ and $SSIM$; Table~\ref{tab3}) drop noticeably, underscoring SCPM’s role in improving fusion quality.
\subsubsection{Analysis of GAM}
To prevent the direct introduction of modality-specific information, GAM modulates the fused features after SCPM processing from a geometric perspective. Ablation experiments show that removing GAM results in fused features lacking modality guidance, leading to insufficient cross-modal interaction. As shown in Fig.~\ref{fig5}, although the model retains some generalisation ability, black background information in SPECT is incorrectly fused. Similarly, the metrics in Tab.~\ref{tab3} also decrease.

\subsubsection{Analysis of TCPM}
TCPM introduces channel perturbation during the decoding stage to help the model adapt to dynamic feature variations. As shown in Fig.~\ref{fig5} and Tab.~\ref{tab3}, removing TCPM leads to a significant decline in performance on medical images, with fusion results incorporating incorrect modality information, thereby reducing overall quality.
\subsubsection{Analysis of Channel Attention}
We introduce the channel attention module (CA) in both SCPM and TCPM. In SCPM, CA is used to compress channels and extract salient features to construct the initial fusion results, while in TCPM, it filters out low-significance channels to prevent redundancy from affecting subsequent reconstruction. The ablation results in Fig.~\ref{fig5} and Tab.~\ref{tab3} show that the removal of CA leads to texture blurring, and the detail expression is weakened.
\subsubsection{Analysis of Pre-trained Model}
In SCPM, we introduce the pre-trained ConvNeXt model to provide global semantic guidance. Ablation experiments show that removing ConvNeXt significantly reduces cross-task performance. This is because the CA module in SCPM relies on training data from a specific modality, and without the guidance of pre-trained knowledge, it may incorrectly select channels when generalising to unseen tasks.


\subsection{Downstream Task Experiments}
We conducted experiments on two downstream tasks: object detection and semantic segmentation. Specifically, BANet \cite{r39} was employed for semantic segmentation on the MSRS dataset, and YOLOv7 \cite{r40} was used for object detection on the M3FD dataset. All fusion methods were evaluated under the same experimental settings. As shown in Tab.~\ref{tab4}, the proposed algorithm achieves the best segmentation accuracy and mIoU across multiple categories. Tab.~\ref{tab5} further shows that our method also leads in two key detection metrics, mAP@0.5 and mAP@[0.5:0.95], confirming its superior fusion performance.


\section{Conclusion}
In this paper, we propose a multi-modality image fusion method based on channel perturbation and pre-training knowledge. Specifically, we design the SCPM to preserve salient features by combining channel attention with global semantic guidance. The GAM is introduced to mitigate the loss of generalisation caused by directly injecting modality information into the encoder. Finally, the TCPM is applied in the decoding stage to improve adaptability by rearranging feature channels. Extensive experiments on fusion tasks and downstream applications validate the effectiveness of the proposed method.

\section{Acknowledgements}
This research was supported by the National Natural Science Foundation of China (No. 62201149), the Basic and Applied Basic Research of Guangdong Province (No. 2023A1515140077), the Natural Science Foundation of Guangdong Province (No.2024A1515011880), the Research Fund of Guangdong-Hong Kong-Macao Joint Laboratory for Intelligent Micro-Nano Optoelectronic Technology (No. 2020B1212030010).

\bibliography{aaai2026}

\end{document}